\documentclass[5pt]{article}

\usepackage[letterpaper]{geometry}
\usepackage{spconf,amsmath,epsfig}
\usepackage{enumitem}
\usepackage{bm}

\usepackage{fancyhdr}
\begin{document}\sloppy

% Example definitions.
% --------------------
\def\x{{\mathbf x}}
\def\L{{\cal L}}

% Title.
% ------
\title{Skeleton-based Action Recognition with Convolutional Neural Networks}
%
% Single address.
% ---------------
\name{Chao Li, Qiaoyong Zhong, Di Xie, Shiliang Pu}
\address{Hikvision Research Institute, Hangzhou, China\\
\{lichao15, zhongqiaoyong, xiedi, pushiliang\}@hikvision.com}
%
% For example:
% ------------
%\address{School\\
%	Department\\
%	Address\\
%   Email}
%
% Two addresses (uncomment and modify for two-address case).
% ----------------------------------------------------------
%\twoauthors
%  {A. Author-one, B. Author-two\sthanks{Thanks to XYZ agency for funding.}}
%	{School A-B\\
%	Department A-B\\
%	Address A-B}
%  {C. Author-three, D. Author-four\sthanks{The fourth author performed the work
%	while at ...}}
%	{School C-D\\
%	Department C-D\\
%	Address C-D\\
%   Email}
%

\maketitle

\begin{abstract}
  Current state-of-the-art approaches to skeleton-based action recognition are
  mostly based on recurrent neural networks (RNN). In this paper, we propose a
  novel convolutional neural networks (CNN) based framework for both action
  classification and detection. Raw skeleton coordinates as well as skeleton
  motion are fed directly into CNN for label prediction. A novel skeleton
  transformer module is designed to rearrange and select important skeleton
  joints automatically. With a simple 7-layer network, we obtain 89.3\%
  accuracy on validation set of the NTU RGB+D dataset. For action detection in
  untrimmed videos, we develop a window proposal network to extract temporal
  segment proposals, which are further classified within the same network. On
  the recent PKU-MMD dataset, we achieve 93.7\% mAP, surpassing the baseline
  by a large margin.
\end{abstract}
\section{Introduction}
\label{sec:intro}

Articulated human pose, also referred to as skeleton, captures full
information needed to understand the underlying activity of the subject.
Compared with other modalities (e.g. RGB images, depth maps), skeleton data
are more robust to noise like background and irrelevant objects. With the
development of low-cost human skeleton capture systems (e.g. Kinect),
large-scale 3D skeleton datasets have been made
available~\cite{shahroudy2016ntu,liu2017pku}, which attract many research
efforts~\cite{han2017space} on skeleton-based human action recognition and
detection. An ever-increasing use of skeleton data in a wide range of
applications from human-computer interaction, virtual reality to video
surveillance can be expected.

Considering the time series property of skeleton sequences in videos,
recurrent neural networks (RNN), in particular long-short term memory networks
(LSTM) are natural choices. Indeed the current state-of-the-art approaches are
mostly based on LSTM. In this paper, we propose a new representation of
skeleton data with convolutional neural networks (CNN), which is shown to
outperform a strong LSTM baseline. Besides, we adapt the widely used Faster
R-CNN~\cite{ren2015faster} object detection framework to action detection in
temporal domain. With the novel CNN-based detection framework, we obtain 58\%
\emph{absolute} mAP improvement over the baseline.

\section{Related Works}

\subsection{Representation of skeleton}

LSTM has been well exploited to model the temporal pattern of skeleton
sequences. Within the LSTM framework, many improvements have been made in the
literature. For example, \cite{zhu2016co} explored the co-occurrence feature
of skeleton joints. \cite{song2017end} exploited attention model in both
spatial and temporal domain. Recently, \cite{zhang2017view} developed a view
adaptive RNN to cope with the viewpoint variations explicitly. On the other
hand, Ke et al.~\cite{ke2017new} proposed a CNN based representation of
skeleton and achieved state-of-the-art performance. Our CNN representation
differs from~\cite{ke2017new} on the input form of skeleton as well as the
network architecture. Besides, we get significantly superior performance
over~\cite{ke2017new}.

\subsection{CNN for object detection}

CNN has achieved great success in many image recognition tasks, e.g. image
recognition, object detection. For object detection, Faster
R-CNN~\cite{ren2015faster} is the current state of the art. It consists of two
cascaded stages. In stage 1, a fully convolutional region proposal network
(RPN) is utilized to extract putative region proposals. In stage 2, features
of the proposals are ROI-pooled and further classified with R-CNN. By sharing
features between RPN and R-CNN, real-time detection is achieved.

Faster R-CNN was originally designed for object detection in still images.
\cite{xu2017r} adapted the framework to temporal activity detection in RGB
videos with a 3D convolutional network. In this work, we are the first to
adapt Faster R-CNN to the task of skeleton-based temporal action detection.

\section{Method}

\subsection{Action classification}

\begin{figure*}[t]
  \centering
  \includegraphics[width=0.65\textwidth]{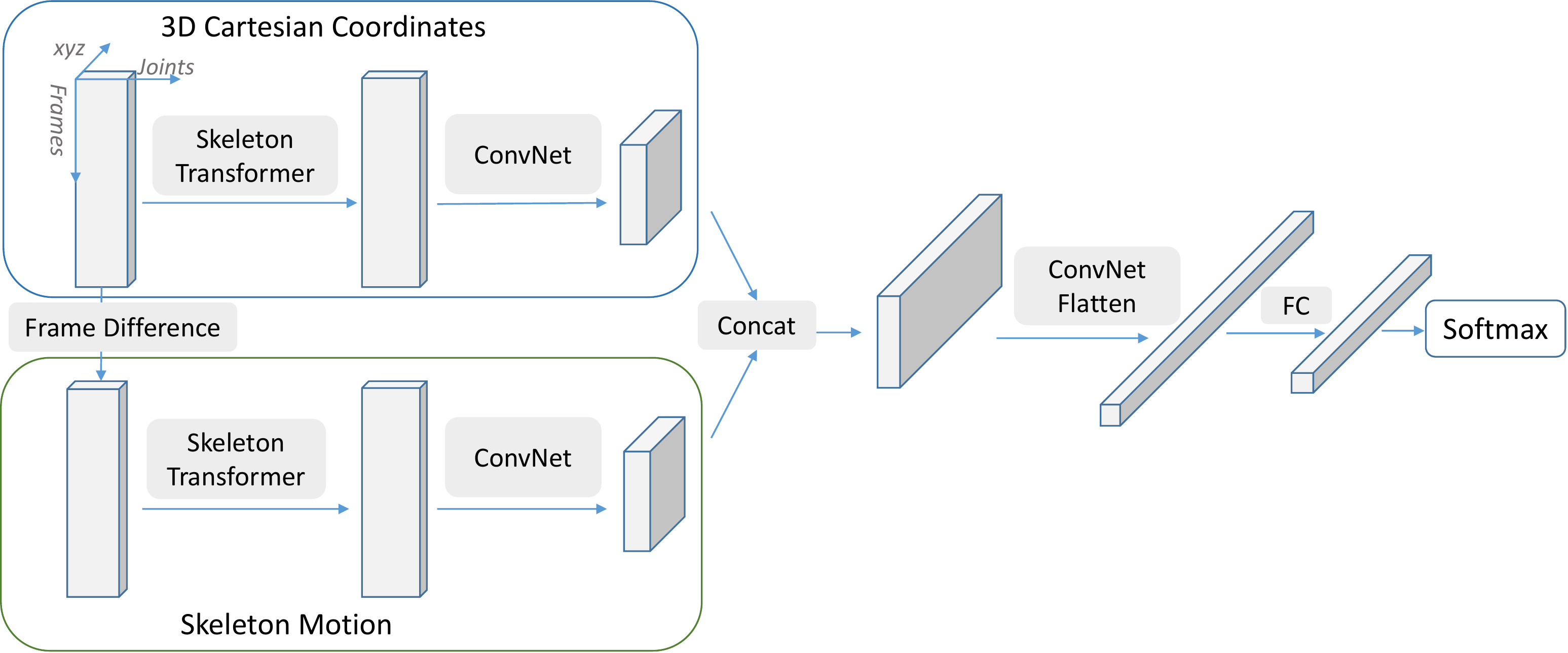}
  \caption{CNN representation of skeleton sequences for action
  classification.}
  \label{fig:act-cls}
\end{figure*}

We design a simple yet effective convolutional architecture
(Fig.~\ref{fig:act-cls}) for action classification from trimmed skeleton
sequences. Besides raw joint coordinates, motion of skeleton joints from two
consecutive frames are fed as an extra input to the network. In addition, we
propose a novel skeleton transformer module. With the module, the network is
able to automatically learn a better ordering of joints as well as new joints
that are more informative than arbitrarily given ones. To deal with
multi-person settings, we use maxout to merge features from skeletons of
different individuals.

\subsubsection{Two-stream CNN}

Two-stream architecture was first introduced in~\cite{simonyan2014two}, where
RGB images and optical flow fields are utilized as two input streams of a
network. Similarly, we define two network inputs for the case of skeleton
data. Given a 3D joint coordinate $\bm{J} = (x, y, z)$, skeleton of one person
is represented as a set of joint coordinates $\bm{S} = \{\bm{J}_1, \bm{J}_2,
\dots, \bm{J}_N\}$ where $N$ is the number of joints per skeleton. Skeleton
motion between two consecutive frames is computed as $\bm{M} = \bm{S}^{t+1} -
\bm{S}^{t} = \{ \bm{J}^{t+1}_1 - \bm{J}^{t}_1, \bm{J}^{t+1}_2 - \bm{J}^{t}_2,
\dots, \bm{J}^{t+1}_N - \bm{J}^{t}_N \}$ where $t$ is frame index. A skeleton
sequence of $T$ frames can be represented as a $T \times N \times 3$ array,
which is treated as a $T \times N$ sized 3-channel image. Raw skeleton
coordinates $\bm{S}$ and skeleton motion $\bm{M}$ are used as two input
streams of our network. Note that we do not perform any normalization.

Actions in a video may span varying length of frames. Since skeleton data are
treated as image data, we normalize all videos to a fixed length by a simple
image resizing operation.

\subsubsection{Skeleton transformer}

For an image, the semantic continuity of pixels is critical. For example, if
we shuffle the location of all pixels randomly, the resulting image would be
non-sense and difficult to recognize for both humans and machines. For the $T
\times N \times 3$ skeleton image data, ordering of $N$ joints are arbitrarily
chosen (e.g. left eye, right eye, nose, \dots), which may not be optimal. To
address this issue, we propose a skeleton transformer module. Given an $N
\times 3$ skeleton $\bm{S}$, we perform a linear transformation $\bm{S}' =
(\bm{S}^T \cdot \bm{W})^T$, where $\bm{W}$ is an $N \times M$ weight matrix.
$\bm{S}'$ is a list of $M$ new interpolated joints. Note that both ordering
and location of the joints are rearranged. The network selects important body
joints automatically, which can be interpreted as a simple variant of
attention mechanism.

Skeleton transformer can be implemented simply with a fully connected layer
(without bias). We place this module at the very beginning of the network
before convolution layers such that it is trained end to end.

\subsubsection{Multi-person maxout}

The methods mentioned above are designed for the case of single person. For
those activities involving human-human interaction (e.g. hugging, shaking
hands), there will be multiple people. A common choice in the literature is to
concatenate skeletons of different people as the network input. Zero padding is
required to deal with varying number of people.

In this paper, we adopt the maxout~\cite{goodfellow2013maxout} scheme for
multiple people. Skeletons of different people go through the same network
layers, and their feature maps are merged by an element-wise maximum operation
after the last convolution layer. The advantage is two-fold. Firstly, the
varying number of people issue can be resolved gracefully without zero
padding. Secondly, by weight sharing, our method can be extended from two
people to more people without increasing model size.

\subsubsection{Network architecture}

We design a tiny 7-layer network which consists of 3 convolution layers and 4
fully connected layers (at which point performance saturates). Our network
contains only 1.3 million parameters. And it can be easily trained from
scratch without any pre-training. Compared with~\cite{ke2017new}, where an
ImageNet pre-trained VGG19 net is used, our model is superior on its compact
model size and fast inference speed as well.

\subsection{Action detection}

\begin{figure*}[t]
  \centering
  \includegraphics[width=0.65\textwidth]{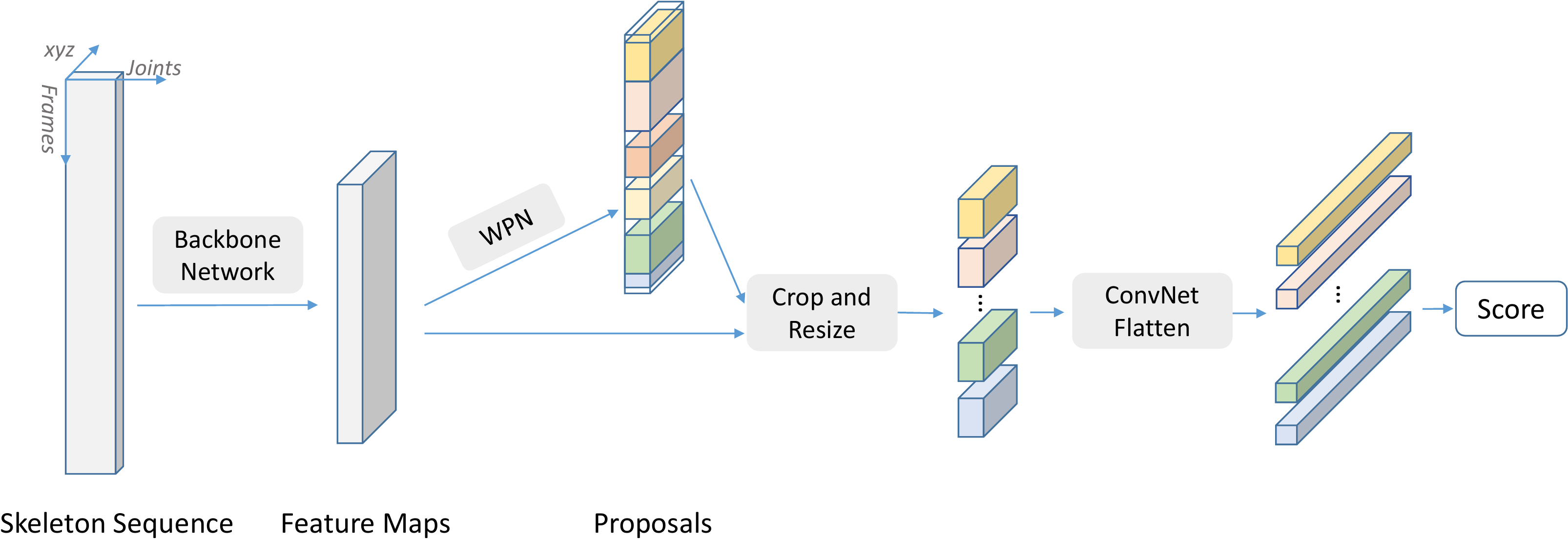}
  \caption{Skeleton-based temporal action detection pipeline.}
  \label{fig:act-det}
\end{figure*}

By interpreting a sequence of skeleton data as a $T \times N \times 3$ image,
it is straightforward to adapt object detection methods to the task of action
detection in temporal domain. In this paper, we take Faster R-CNN as an
example. Other object detection frameworks should work as well.

As displayed in Fig.~\ref{fig:act-det}, the region proposal network (RPN) is
replaced with a window proposal network (WPN) . In particular, 2D anchors are
flattened to 1D anchors. Window proposals along the temporal dimension are
extracted based on pre-defined anchors. Window regression instead of bounding
box regression is performed to refine the temporal position of window
proposals. After the proposals are ready, we pool features of each window from
the shared feature maps with the crop-and-resize operation. These features are
then fed to the R-CNN subnetwork for classification and window regression. For
the backbone network, we use the same architecuture as action classification
in Fig.~\ref{fig:act-cls}.

In our experiments, we use 4 anchor scales, i.e. \{50, 100, 200, 400\}. For
other hyper-parameters, we follow the settings recommended
in~\cite{ren2015faster}. During training, we randomly choose a temporal scale
factor between 0.8 and 1.5. During testing, we use single scale (the original
resolution).

\section{Experiments}

We validate our method on two large-scale skeleton datasets. The NTU RGB+D
dataset~\cite{shahroudy2016ntu} is designed for action classification task. It
contains 56880 well trimmed video clips spanning 60 action categories. The
very recent PKU-MMD dataset~\cite{liu2017pku} is designed for action detection
task, which contains 1076 untrimmed videos and 21545 action instances. The
number of action categories is 51.

\subsection{Action classification}

\subsubsection{Ablation study}

To evaluate contributions of different components, we perform an ablation
study on the NTU RGB+D dataset. Table~\ref{tab:act-cls-abl} shows the results.
Using plain CNN, we already outperform STA-LSTM~\cite{song2017end} (see
Table~\ref{tab:act-cls}), a strong LSTM baseline. Skeleton motion improves
accuracy by 1.6 and 3.3 points in cross-subject and cross-view settings
respectively. Skeleton transformer improves cross-subject by 1.8 points, while
improvement on cross-view is marginal. This could be explained by the reason
that variation of actions across subjects is larger than that of across views,
which can be alleviated by skeleton transformer. Combining skeleton motion and
skeleton transformer, we obtain 83.2\% and 89.3\% accuracy in the two
partitioning schemes respectively.

\begin{table}[t]
\begin{center}
\caption{Ablation study on action classification. Accuracies are measured on
  validation set of the NTU RGB+D dataset.}
\label{tab:act-cls-abl}
\begin{tabular}{|c|c|c|}
  \hline
  Method & Cross-subject & Cross-view \\
  \hline
  CNN & 0.798 & 0.852 \\
  \hline
  CNN+Motion & 0.814 & 0.885 \\
  \hline
  CNN+Trans & 0.816 & 0.854 \\
  \hline
  CNN+Motion+Trans & \textbf{0.832} & \textbf{0.893} \\
  \hline
\end{tabular}
\end{center}
\end{table}

\subsubsection{Comparison to the state-of-the-arts}

Our method significantly outperforms all recent state-of-the-art approaches in
both cross-subject and cross-view settings (Table~\ref{tab:act-cls}).
Specifically, we improve accuracy by 10 points over STA-LSTM in cross-subject
setting. Besides, our method is also superior over Ke et al.~\cite{ke2017new},
which is also based on CNN. The excellent result clearly proves the ability of
CNN to model temporal pattern. We believe that CNN can be applied to other
time series signals other than skeleton sequences.

\begin{table}[t]
\begin{center}
\caption{Action classification performance on validation set of the NTU RGB+D
  dataset.}
\label{tab:act-cls}
\begin{tabular}{|c|c|c|}
  \hline
  Method & Cross-subject & Cross-view \\
  \hline
  STA-LSTM~\cite{song2017end} & 0.734 & 0.812 \\
  \hline
  VA-LSTM~\cite{zhang2017view} & 0.792 & 0.877 \\
  \hline
  Ke et al.~\cite{ke2017new} & 0.796 & 0.848 \\
  \hline
  Proposed & \textbf{0.832} & \textbf{0.893} \\
  \hline
\end{tabular}
\end{center}
\end{table}

\subsection{Action detection}

We validate our action detection pipeline on the PKU-MMD dataset. Since the
official evaluation code is unavailable, we report our performance based on
our own implementation of mean average precision (mAP) over different actions.
Table~\ref{tab:act-det} shows the mAP numbers at two different Intersection
over Union (IoU) thresholds. The detector easily enjoys benefits from the
capability of our CNN-based classifier. Compared with the strong baseline
JCRRNN~\cite{li2016online}, we obtain a performance boost. That is 58\%
\emph{absolute} mAP improvement in cross-subject and 40\% mAP improvement in
cross-view at IoU threshold of 0.5. The performance improvement indicates that
it is a viable solution to treat skeleton sequences as images and transform
the temporal action detection problem into a unidimensional object detection
problem.

\begin{table}[t]
\begin{center}
\caption{Action detection performance on validation set of the PKU-MMD
  dataset. $\theta$ is the IoU threshold.}
\label{tab:act-det}
\begin{tabular}{|c||c|c||c|c|}
  \hline
  Partition & \multicolumn{2}{c||}{Cross-subject} & \multicolumn{2}{c|}{Cross-view} \\
  \hline
  $\theta$ & 0.1 & 0.5 & 0.1 & 0.5 \\
  \hline
  JCRRNN~\cite{li2016online} & 0.452 & 0.325 & 0.699 & 0.533 \\
  \hline
  Proposed & \textbf{0.922} & \textbf{0.904} & \textbf{0.958} & \textbf{0.937} \\
  \hline
\end{tabular}
\end{center}
\end{table}

\section{Conclusion}

Skeleton based human action recognition is drawing more and more attention as
the popularity of 3D skeleton data. By treating skeleton sequences as images,
we propose a novel CNN based framework for both action classification and
detection tasks. Our method achieves new state-of-the-art performance on two
recent large-scale skeleton datasets. The proposed action detection approach
detects actions in a batch-processing way, while online detection is required
for real-time applications. We leave it as future exploration.

\bibliographystyle{IEEEbib}
\bibliography{main}

\end{document}